\documentclass[10pt,twocolumn,letterpaper]{article}

\usepackage{wacv}
\usepackage{times}
\usepackage{epsfig}
\usepackage{graphicx}
\usepackage{amsmath}
\usepackage{amssymb}

\usepackage{mathtools}
\usepackage{setspace}
\usepackage{algorithm}
\usepackage[noend]{algpseudocode}

\usepackage{slashbox}
\usepackage{ruler}
\usepackage{booktabs}       
\usepackage{color}
\usepackage{dcolumn}
\usepackage{multirow}

\usepackage[pagebackref=true,breaklinks=true,letterpaper=true,colorlinks,bookmarks=false]{hyperref}

\wacvfinalcopy 

\def\wacvPaperID{927} 

\begin{document}

\title{Generative Pseudo-label Refinement for Unsupervised Domain Adaptation}
\author{Pietro Morerio\textsuperscript{1}, Riccardo Volpi\textsuperscript{1}, Ruggero Ragonesi\textsuperscript{1}, Vittorio Murino\textsuperscript{1,2,3}\\
{\tt\small \{pietro.morerio,riccardo.volpi,ruggero.ragonesi,vittorio.murino\}@iit.it}\\~\\
\textsuperscript{1}Pattern Analysis \& Computer Vision - Istituto Italiano di Tecnologia\\
\textsuperscript{2}Computer Science Department - Universit\`a di Verona, Italy\\
\textsuperscript{3}Huawei Technologies Ltd., Ireland Research Center
}

\maketitle

\begin{abstract}
We investigate and characterize the inherent resilience of conditional Generative Adversarial Networks (cGANs) against noise in their conditioning labels, and exploit this fact in the context of Unsupervised Domain Adaptation (UDA). In UDA, a classifier trained on the labelled source set can be used to infer pseudo-labels on the unlabelled target set. However, this will result in a significant amount of misclassified examples (due to the well-known domain shift issue), which can be interpreted as noise injection in the ground-truth labels for the target set. We show that cGANs are, to some extent, robust against such ``shift noise''. Indeed, cGANs trained with noisy pseudo-labels, are able to filter such noise and generate cleaner target samples. We exploit this finding in an iterative procedure where a generative model and a classifier are jointly trained: in turn, the generator allows to sample cleaner data from the target distribution, and the classifier allows to associate better labels to target samples, progressively refining target pseudo-labels. Results on common benchmarks show that our method performs better or comparably with the unsupervised domain adaptation state of the art.





\end{abstract}

\section{Introduction}\label{intro}

Unsupervised Domain Adaptation (UDA) addresses the problem of learning models that perform well on a \textit{target} domain for which ground truth annotations are not provided. During the training phase, one can leverage unlabeled samples from this distribution and labelled samples from a \textit{source} distribution, separated by the so-called \textit{domain shift} \cite{NameTheDataset}, \ie, drawn from two different data distributions. 
In this work, we address UDA from a novel perspective, by casting the problem in the setting of \textit{learning with noisy labels}~\cite{Rolnick2017DeepLI}. 

\begin{figure}[!t]
\centering
\includegraphics[width=\linewidth]{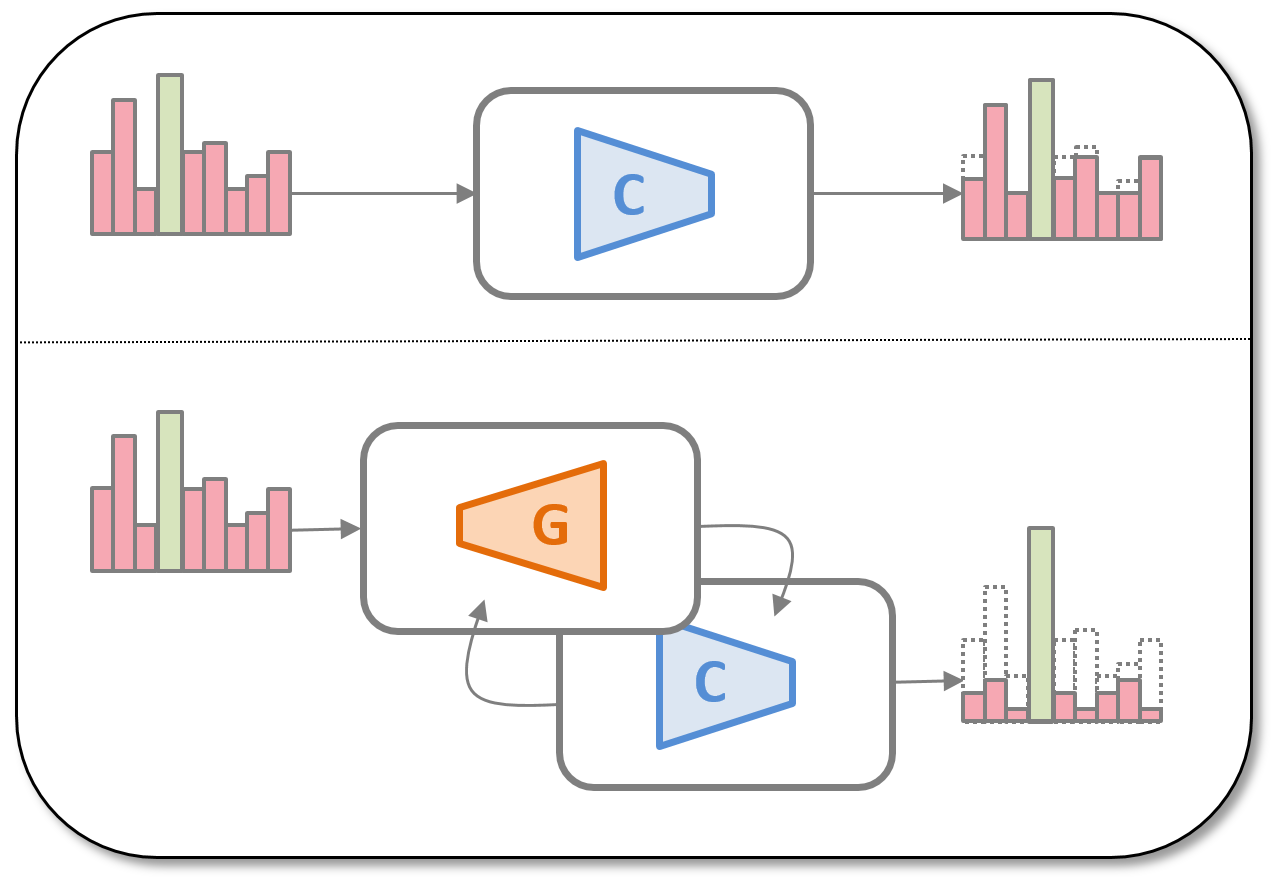}
\caption{\footnotesize \textit{Top}: a neural network classifier $C$ is typically not robust against shift noise, here represented by an histogram.
\textit{Bottom}: a cGAN generator $G$ is able to filter such structured noise, making it more uniform and thus tolerable from the classifier. By jointly training  $C$ and $G$, the former benefits from ``cleaner" generated data, while the latter from more accurate inferred labels.}
\vspace{-5pt}\label{fig:simple}
\end{figure}

We start from the very simple realization that, given a model trained on the source domain, we can infer a set of (pseudo-)labels for the target domain. Typically, due to the domain shift, a consistent number of labels are wrongly inferred, resulting in a noisy-labelled target set, with an amount of noise proportional to the classification error. Previous work~\cite{Rolnick2017DeepLI} has shown that deep learning-based classifiers are robust against label noise, provided a sufficiently large training set. For this reason, one might hope that such resilience could be exploited to train a classifier for the target domain with the supervision of the inferred, noisy pseudo-labels. The idea is that the classifier might disregard label noise to some extent, providing target accuracy higher than the original model trained on source samples. 

However, we empirically show that such strategies alone cannot compete with existing UDA methods in terms of accuracy on the target set. Indeed, while deep models' robustness against noisy labels is remarkable when the noise distribution is nearly uniform, we show that they are not robust against the label noise resulting from the domain shift. Although \textit{a priori} assumptions cannot be made on such noise, we empirically observe in a variety of adaptation problems heavy deviations from uniform noise, meaning that misclassications are not evenly distributed across classes. We term such highly structured noise \textit{ ``shift noise''}. 

While classifiers are not robust against such more structured kind on noise, we observe that conditional Generative Adversarial Networks~\cite{CGAN} (cGAN) are. A cGAN model can be trained to generate samples conditioning on the desired classes of an arbitrary distribution. It was shown that this class of models can be \textit{made} more resistant against label noise~\cite{NIPS2018_8229}, but we provide empirical evidence that---to some extent---they are inherently robust against it, without any modification from the standard training procedure~\cite{GAN,CGAN}. This means, in practice, that training a cGAN on some noisy-labelled samples will result in a model that generates samples that are ``cleaner'' than the training ones. A natural idea that follows this finding, is trying to generate cleaner target samples to train better performing models for the target domain. However, although cGANs are to some extent resistant to noisy-labels, they are not robust enough to generate target samples that allow to train competitive models. 

Interestingly though, we observe that, even if the noise reduction is not sufficient to train competitive target models, \textit{the labels of the generated samples obey a noise distribution which is closer to the uniform than the shift noise one}.

In this work, we explore the two facts above (classifier robustness against uniform noise~\cite{Rolnick2017DeepLI} and cGAN robustness to shift noise), and jointly exploit them for UDA. We devise a UDA strategy based on the properties of both classifiers and cGANs to filter out noise in the labels. We propose an iterative procedure, where we alternately optimize the losses associated with a cGAN and with a classifier. Throughout the training phase, the classifier can benefit from more and more reliable conditionally-generated data, while a cGAN can exploit more and more reliable pseudo-labels inferred by the classifier (see Figure~\ref{fig:simple}). Source samples are only exploited to train an initial classifier. After this step, the problem is faced in a fully unsupervised fashion, reducing the noise on the labels of the empirical target distribution over iterations during training. Results on standard UDA benchmarks show the effectiveness of our approach.

\textbf{Summary.} The main contributions of this work are the following: 
\textbf{(i)} we characterize the concept of shift noise, and provide an analysis of the robustness of both discriminative and generative models against it;
\textbf{(ii)} we design a novel training procedure that leverages the above findings in order to refine the predictions of a classifier over iterations; 
\textbf{(iii)} we apply the proposed algorithm in the unsupervised domain adaptation scenario, observing competitive performance with the state of the art on public benchmarks. 

The remaining of the paper is organized as follows. In Section~\ref{sec:rel_work}, we detail background and related work. In Section~\ref{sec:robustness}, we characterize shift noise and investigate the robustness of classifiers and cGANs against it.  Section~\ref{sec:uda} describes how to exploit these findings to tackle UDA problems, and Section~\ref{sec:exp} reports the related experimental results. Finally, we draw the conclusions in Section~\ref{sec:concl}.

\section{Background and related work}
\label{sec:rel_work}


\paragraph{Unsupervised Domain Adaptation.} 

In UDA, we are given a set of samples from a source distribution in the form $\{x_s,y_s\} \sim p_{source}$, and a set of samples from a target distribution of interest in the form $\{x_t\} \sim p_{target}$ (no labels). The goal is to perform well on data from the target distribution. Different approaches allow to solve this problem efficiently in a plethora of tasks. Adversarial training has been effectively used to map source and target samples in a common feature space \cite{Ganin,Ganin2,ADDA,volpi2018cvpr}. Other works aim at aligning the second order statistics of source and target features \cite{DeepCORAL,morerio2018}. More recently, image-to-image translation methods, that learn the mapping from the source space to the target one and vice-versa, have been proposed \cite{CoGAN,DTN,Russo_2018_CVPR,GOOGLE,UITITN,GenToAdapt,CycADA}. In general, one can design models for UDA that leverage labeled source samples that are ``rendered" with the style of target samples (and vice-versa). Other works propose different successful solutions to face the adaptation problem (\emph{e.g.}, \cite{DSN,Tri,Transductive,ADA,saito2018adversarial,Saito_2018_CVPR}). Since the latter are only related to our work for the common goal, they are not detailed in this section. 
Our approach is somehow related to image-to-image translation methods, since we exploit generated samples to train a classifier for the target domain. In particular, PixelDA~\cite{GOOGLE} is the most related method, since it leverages a training procedure where a GAN and a classifier are jointly trained. However, the latter makes a strong assumption on the relationship between source and target domains: \textit{``the differences between the domains are primarily low-level (due to noise, resolution, illumination, color) rather than high-level (types of objects, geometric variations, etc)''}. In this work, we generate target images using a simple cGAN, namely mapping noise vectors from a latent space into the image space, merely conditioning on label codes. This difference comes with two main advantages: our architecture and loss functions are much simpler than the ones adopted for image-to-image translation, and we do not have to make such strong assumptions on the gap between the two domains. 

Our method is substantially different from most UDA solutions also because we do not need source samples throughout the adaptation procedure, but only to pre-train the model $M_{\theta_s}$, used to assign pseudo-labels to target samples. Indeed, solutions that align source/target feature statistics \cite{DeepCORAL,morerio2018}, map samples from both distributions in a common feature space via adversarial training \cite{Ganin,Ganin2}, or translate images between domains \cite{CoGAN,DTN,Russo_2018_CVPR,GOOGLE,UITITN,GenToAdapt,CycADA}, are typically based on objectives that depend on both source and target samples. 
In our case, the independence from source samples during the adaptation procedure brings a number of advantages. The main one is that the training procedure designed for a certain target can be used \emph{as is}, regardless of the source domain, the only difference being the model $M_{\theta_s}$ used for the first, initial label inference. Moreover, many adaptation methods require additional hyperparameters to balance different loss terms \cite{Ganin,Ganin2,UITITN,DTN,Russo_2018_CVPR,GenToAdapt,GOOGLE} that depend on both source and target samples. The latter is a huge drawback because in UDA we do not have target labels for hyperparameter cross-validation.


\paragraph{Learning with pseudo-labels.} 
Our joint training procedure for UDA is related to the approach by Lee et al. \cite{Lee_pseudo_label}. In this work, a method for semi-supervised learning is proposed, where, as training proceeds, inference is performed on unlabeled samples, and the pseudo-labels obtained are interpreted as correct and used for training a classifier. Part of our method has similarities to this idea since, during our training procedure, we infer pseudo-labels for the target samples. However, we are different in that we use them to train a generative model.

\paragraph{Generative Adversarial Networks.} 

The original formulation by Goodfellow et al. \cite{GAN} is defined by the following minimax game between a network $D$ (discriminator) and a network $G$ (generator)

\begin{align}
\vspace{-5pt}
\min_{\theta_{D}} \max_{\theta_{G}} \: &\mathcal{L}_{GAN} = \mathbb{E}_{x \sim p_{x}} [-\log D(x;\theta_D)] \\ 
&\quad+ \mathbb{E}_{z \sim p_{z}} [-\log(1 - D(G(z;\theta_G);\theta_D ))] \nonumber
\vspace{-5pt}
\end{align}


Solving such optimization problem makes $D$ classify samples from the data distribution as \emph{real} and samples generated by $G$ as \emph{fake}. Conversely, it makes $G$ generate samples that $D$ would classify as \emph{real}. 

A straightforward extension is the concatenation of label codes to the input before it is fed to $G$ and $D$, in order to condition on the class from which data are generated. This extension is termed \emph{conditional} GAN (cGAN, \cite{CGAN}), and represents the class of models this work focuses on, being our method based on class-conditioned image generation.

Several alternatives to the original GAN formulation \cite{GAN} have been proposed. Two examples are substituting the cross-entropy loss with the least-squares loss \cite{LSGAN} or with the Hinge loss \cite{miyato2018spectral}. Also more elaborated alternatives have been introduced \cite{WGAN,DRAGAN,BEGAN}. To date, the superiority of one objective function over the others is not fully clear \cite{equal_GANs}, and the main advancements on GAN research have been related to architectural choices \cite{DCGAN} and different training procedures \cite{CycleGAN,miyato2018spectral,karras2018progressive,brock2018large}.

\section{Robustness against label noise}
\label{sec:robustness}

\addtolength{\tabcolsep}{3pt}    
\begin{table*}[h]
\begin{center}
\small
\begin{tabular}{cccc}
\toprule

\textbf{Shift noise} & \textbf{Classifier} &  \textbf{GAN-test \cite{Shmelkov_2018_ECCV}} &  \textbf{GAN-train \cite{Shmelkov_2018_ECCV}}\\
\midrule 
\includegraphics[width=.2\linewidth]{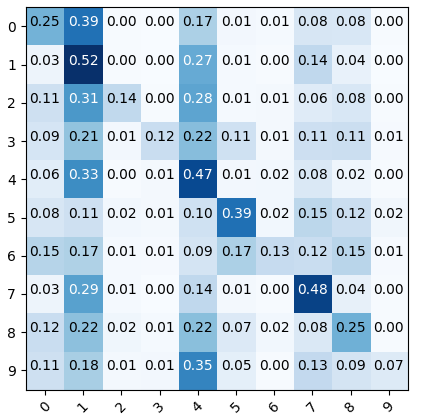}  &
\includegraphics[width=.2\linewidth]{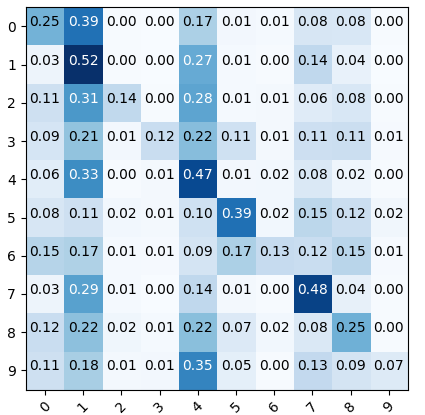} &
\includegraphics[width=.2\linewidth]{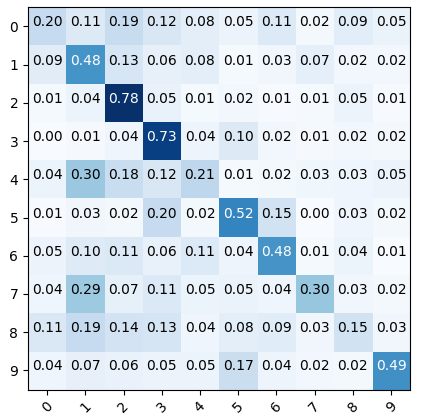} &
\includegraphics[width=.2\linewidth]{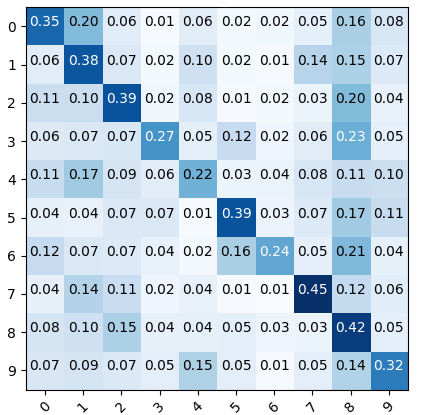} \\

\cmidrule(lr){1-1}  \cmidrule(lr){2-2} \cmidrule(lr){3-3} \cmidrule(lr){4-4}

$a$ = 0.300 & $a$ = 0.321 $\pm$ 0.002 & $a$ = 0.419 $\pm$ 0.012 & $a$ = 0.337 $\pm$ 0.010\\
$\delta_A$ = 0.374 & $\delta_A$ = 0.374 $\pm$ 0.0001 & $\delta_A$ = 0.252 $\pm$ 0.012 & $\delta_A$ = 0.270 $\pm$ 0.006 \\

\bottomrule
\end{tabular}
\end{center}
\vspace{-5pt}
\caption{\textit{Left:} shift noise for the split MNIST$\rightarrow$SVHN: only 30\% of the labels are correctly inferred on SVHN after training a classifier on MNIST; high degree of asymmetry (values refer to the training sets). \textit{Mid-left}: the confusion matrix, accuracy and $\delta_A$ for a classifier trained on noisy SVHN almost reflect the initial ones, meaning that shift noise was overfitted. \textit{Mid-right} Oracle performance on samples generated by a cGAN trained with shift noise. Not only generated images are classified better than training samples, but also residual noise is less structured (lower $\delta_A$). \textit{Right}: A classifier is trained on (cleaner) cGAN-generated samples: its accuracy is slightly higher than the one of the classier directly trained on shift noise, but more importantly inferred labels show a consistently lower amount of structure in their noise. Results are averages over 5 runs, starting from the same shift noise and accuracies refers to the training sets.}
\label{tab:confmats_small}
\end{table*}
\addtolength{\tabcolsep}{-3pt}

In this section we first formalize the problem and describe the concept of \textit{shift noise}, as the noise resulting from inferring labels in a domain that is different from the training one. Armed with the formal definition, we explore the robustness of ConvNets~\cite{ConvNet} and cGANs against such peculiar and highly structured noise. Following standardized UDA benchmarks used by the main competing algorithms~\cite{DTN,CoGAN,GOOGLE,UITITN,Russo_2018_CVPR}---namely works that rely on GANs to perform adaptation---we train models on MNIST~\cite{MNIST} and test on SVHN~\cite{SVHN}, MNIST-M~\cite{Ganin} and USPS~\cite{USPS}, we train on SVHN and test on MNIST, and we train on USPS and test on MNIST. For brevity, we define the procedure of training on source and testing on target as $source \rightarrow target$ (\eg, MNIST $\rightarrow$ SVHN). The conclusions we draw in this section will motivate the algorithmic choices we will introduce in Sec. \ref{sec:uda}





\begin{table*}[h]
\begin{center}
\small{
\begin{tabular}{rcccccccc}
\toprule
 & \multicolumn{2}{c}{\textbf{Shift noise}} &  \multicolumn{2}{c}{\textbf{Classifier}} &  \multicolumn{2}{c}{\textbf{Equiv. unif. noise} } & \multicolumn{2}{c}{\textbf{Classifier}} \\

\cmidrule(lr){2-3}  \cmidrule(lr){4-5} \cmidrule(lr){6-7} \cmidrule(lr){8-9} 

\textbf{Split} & $a$ & $\delta_A$ & $a$ & $\delta_A$ & $a$ & $\delta_A$ & $a$ & $\delta_A$ \\
\midrule

SVHN $\rightarrow$ MNIST  
& 0.669             & 0.208 
& 0.660	$\pm$ 0.001 & 0.241 $\pm$ 0.003 
& 0.669             & 0.017
& 0.879	$\pm$ 0.043 & 0.029 $\pm$	0.006\\

MNIST $\rightarrow$ SVHN 
& 0.300             & 0.374 
& 0.321	$\pm$ 0.002 & 0.374 $\pm$ 0.000 
& 0.300             & 0.157
& 0.293	$\pm$ 0.006 & 0.161 $\pm$	0.008\\

MNIST $\rightarrow$ MNIST-M 
& 0.550             & 0.153 
& 0.557 $\pm$ 0.003 & 0.153 $\pm$ 0.000 
& 0.550             & 0.014
& 0.619 $\pm$ 0.002 & 0.023 $\pm$ 0.001\\

USPS $\rightarrow$ MNIST 
& 0.608             & 0.273
& 0.619	$\pm$ 0.001 & 0.273 $\pm$ 0.000 
& 0.608             & 0.013
& 0.881	$\pm$ 0.036 & 0.026 $\pm$	0.006 \\

MNIST $\rightarrow$ USPS 
& 0.819             & 0.150 
& 0.807	$\pm$ 0.002 & 0.150 $\pm$ 0.000
& 0.819             & 0.024
& 0.919 $\pm$ 0.006 & 0.025 $\pm$ 0.001\\
\bottomrule
\end{tabular}
} 
\end{center}
\vspace{-5pt}
\caption{Classifiers tend to overfit shift noise, reflecting initial accuracy and asymmetry of the shift noise itself. They are instead significantly robust to an equivalent (in term of number of corrupted labels) amount of uniform noise $n$ (eq. \ref{eq:noise_level}).}
\label{tab:delta_acc_classif}
\end{table*}
\begin{table*}[h]
\begin{center}
\small{
\begin{tabular}{rcccccc}
\toprule
 & \multicolumn{2}{c}{\textbf{Shift noise}} &  \multicolumn{2}{c}{\textbf{GAN-test}} &  \multicolumn{2}{c}{\textbf{GAN-train}} \\

\cmidrule(lr){2-3}  \cmidrule(lr){4-5} \cmidrule(lr){6-7} 

\textbf{Split} & $a$ & $\delta_A$ & $a$ & $\delta_A$ & $a$ & $\delta_A$ \\
\midrule
SVHN $\rightarrow$ MNIST  
& 0.669             & 0.208 
& 0.737 $\pm$ 0.018 & 0.134 $\pm$ 0.006 
& 0.725	$\pm$ 0.014 & 0.219	$\pm$ 0.007 \\

MNIST $\rightarrow$ SVHN 
& 0.300             & 0.374 
& 0.419 $\pm$ 0.012 & 0.252	$\pm$ 0.012 
& 0.337 $\pm$ 0.010 & 0.270	$\pm$ 0.006\\

MNIST $\rightarrow$ MNIST-M 
& 0.550             & 0.153 
& 0.565 $\pm$ 0.057 & 0.189 $\pm$ 0.067  
& 0.536	$\pm$ 0.092 & 0.174	$\pm$ 0.068\\

USPS $\rightarrow$ MNIST 
& 0.608             & 0.273
& 0.772 $\pm$ 0.006 & 0.114 $\pm$ 0.005 
& 0.692	$\pm$ 0.018 & 0.239 $\pm$ 0.009\\

MNIST $\rightarrow$ USPS 
& 0.819             & 0.150 
& 0.810 $\pm$ 0.005 & 0.135 $\pm$ 0.010 
& 0.824	$\pm$ 0.005 & 0.143	$\pm$ 0.004 \\
\bottomrule
\end{tabular}
} 
\end{center}
\vspace{-5pt}
\caption{cGANs trained with shift noise show robustness in generating clean samples (GAN-test), according to an oracle classifier. At the same time, they produce samples with enough variability and quality: in fact a classifier trained on generated samples outperforms a classifier trained directly on shift noise (Table \ref{tab:delta_acc_classif}) both in term of accuracy and noise uniformity. }
\label{tab:delta_acc_GAN}
\end{table*}

\subsection{Shift noise.}\label{subsec:shiftnoise}

In the UDA setting, given a model $M_{\theta_s}(x)$, trained  on a source domain $\mathcal{S}=\{x^{(i)}_s, y^{(i)}_s\}_{i=1}^n$, we can infer a pseudo-label $\Tilde{y}=M_{\theta_s}(x_t)$ for each target sample $x_t$. Mis-classification on the target set will result in a noisy set of pseudo-label associations $\mathcal{T}=\{x^{(i)}_t,\Tilde{y}^{(i)}\}_{i=1}^m$. 

Table \ref{tab:confmats_small} (first column) provides an example, associated with the split MNIST $\rightarrow$ SVHN. First of all, we can observe that misclassification noise in the confusion matrices, which we term \textit{shift noise}, is not uniformly distributed across classes. Moreover, shift noise is also different from the structured noise analyzed by Rolnick et al.~\cite{Rolnick2017DeepLI}, where the correct label is always assumed to have the highest probability. The only \textit{a priori} assumption we can make concerns the amount of shift noise, which must at least guarantee an accuracy higher than random chance for $M_{\theta_s}(x)$ on $\mathcal{T}$. Note that given an accuracy $a$ for $M_{\theta_s}(x_t)$, the same accuracy can be obtained by injecting \textit{uniform} noise in a fraction $n$ of the labels:
\begin{equation}\label{eq:noise_level}
\vspace{-5pt}
    n=(1-a)\frac{c}{c-1},
\end{equation} 
where $c$ is the number of classes. Vice-versa, randomizing a fraction $n$ of the labels, one would get a fraction of correct predictions (i.e. accuracy) equal to 
\begin{equation}\label{eq:accuracy_level}
\vspace{-5pt}
    a = 1 - n \frac{c-1}{c}.
\end{equation} 
Note that randomizing a fraction $n$ of labels does not imply they are all wrong, since, on average, $1/c$ of them will be assigned the correct class.

As already mentioned, no hypotheses on the shape of the distribution of shift noise can be put forward, making it difficult to characterize it. However, a useful estimate of its amount of structure is given by the asymmetry of the confusion matrix $M$, defined as:  
\begin{equation}\label{eq:asymmm}
    \delta_A(M) = \dfrac{||M-M^T||_F}{2||M||_F}.
\end{equation}
In general, $0 \le \delta_A(M) \le 1$. We have $\delta_A(M) = 0 $ for symmetric matrices, thus $\delta_A(M) \approx 0 $ for uniform noise, since $M$ would be approximately symmetric. For shift noise, $ 0 < \delta_A(M) < 1 $. The lower $\delta_A$ the more uniform the noise. 
Values for $\delta_A$ are given for all the considered benchmarks in Table \ref{tab:delta_acc_classif}, together with the amount of correctly inferred labels (\ie accuracy).

\subsection{Classifiers}
\label{subsec:classif}

\begin{figure*}[!t]
\centering
\includegraphics[width=0.8\linewidth]{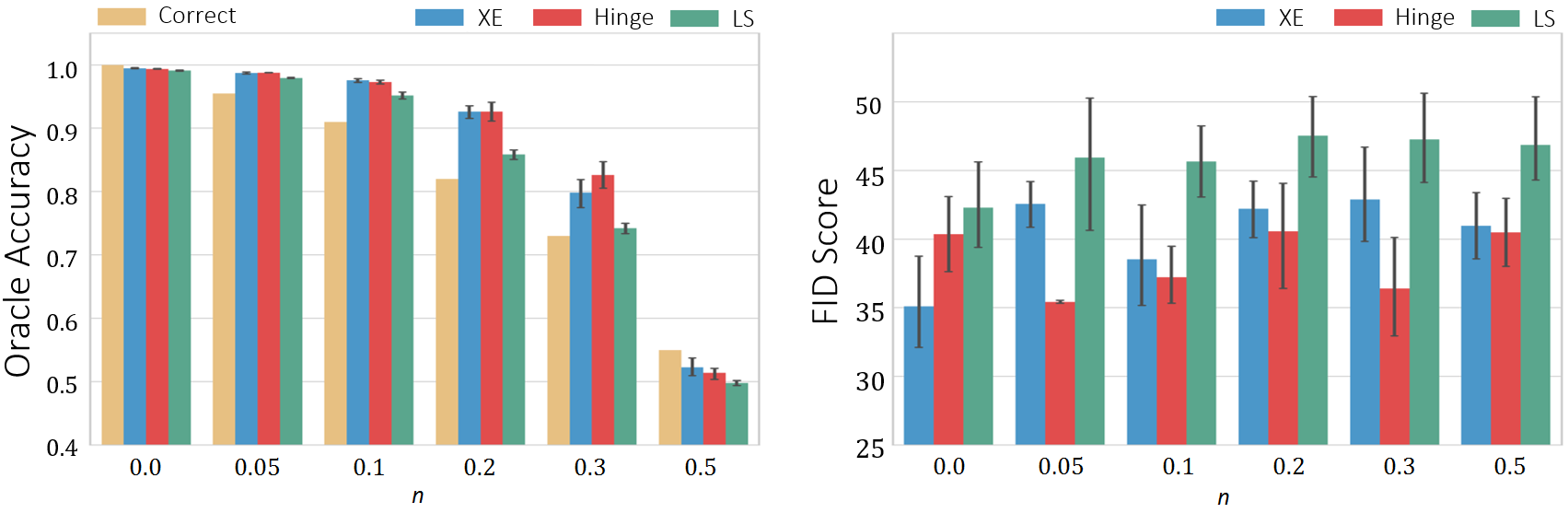}
\vspace{-3pt}
\caption{\footnotesize \emph{Left} panel plots the fraction of images correctly generated by cGANs with different levels $n$ of uniform noise and different objectives (\emph{blue}: cross-entropy \cite{GAN}, \emph{red}: Hinge \cite{miyato2018spectral}, \emph{green}: least-squares \cite{LSGAN}), evaluated through the oracle. \emph{Yellow} bars indicate the percentage of images with the correct label in the training set. \emph{Right} panel shows the FID scores achieved with different levels of noise and different GAN objectives (same as \emph{left}). 
}
\label{fig:noisy_mnist}
\end{figure*}

As already mentioned, a study on the robustness of classifiers against label noise is proposed by Rolnick et al.~\cite{Rolnick2017DeepLI}. We integrate their findings by training a set of classifiers with labels corrupted by shift noise. In practice, we train a first ancillary classifier $M_{\theta_s}(x)$ on a source domain $\mathcal{S}$, and then use it to assign labels $\Tilde{y}=M_{\theta_s}(x_t)$ for the samples of a target domain $\mathcal{T}$. Eventually, we train a new classifier from scratch on the noisy set $\mathcal{T}=\{x^{(i)}_t,\Tilde{y}^{(i)}\}_{i=1\dots m}$, where labels $\Tilde{y}$ are corrupted by shift noise deriving from misclassifications produced by $M_{\theta_s}$.

Classifiers tend to tolerate uniform noise in training labels to a good extent \cite{Rolnick2017DeepLI}.
Differently, as reported in Table \ref{tab:delta_acc_classif} - columns 2 and 3 - we notice that a classifier trained with shift noise in the training labels is perfectly able to (over)fit it, being accuracies on the training set and noise asymmetry in the confusion matrices nearly the same. This can also be noticed when comparing the first and second columns of Table~\ref{tab:confmats_small}.
Note that this behavior does not depend on the \textit{amount} of noise, but only on its nature. In fact, training the same classifiers with the equivalent amount of uniform noise (eq. \ref{eq:noise_level}) produces higher accuracies (together with, of course, nearly null $\delta_A$), as shown in Table \ref{tab:delta_acc_classif}, columns 4 and 5. Note also that we train all classifiers till convergence, since we have no means for early stopping. This reflect the UDA setting were no target validation labels are provided.

Since deeper architectures, and Residual Networks \cite{ResNet} in particular, are more robust against uniform noise \cite{Rolnick2017DeepLI}, one could wonder whether they could prove more resilience against shift noise than shallow models: this does not happen as we show in the experiments provided in the Supplementary Material.

\subsection{cGANs}\label{subsec:cgan}

Given the success of GANs in UDA, we investigate their properties in term of robustness against both uniform and shift noise, since never investigated before.

\paragraph{Uniform noise.}  We consider the MNIST dataset \cite{MNIST} and assume to have an oracle to classify which class a sample belongs to. In practice, this oracle is a ConvNet trained on MNIST, that achieves $>99\%$ accuracy on both the training and the test sets. Accuracy of such oracle on GAN-generated samples is referred to as the \textit{GAN-test} metric \cite{Shmelkov_2018_ECCV}.

One might genuinely expect that training a cGAN with, \emph{e.g.}, a fraction of noisy labels $n=0.1$ will result in $\sim 10\%$ of mis-generated samples. We show in the following that this does not occurr. We train the cGAN with different levels of uniform noise $n$ and evaluate the output of the generator through the oracle, by comparing the label code given in input to the cGAN and the output of the oracle fed with the generated image. 

Figure~\ref{fig:noisy_mnist} \textit{(left)} reports our findings. Yellow bars indicate the percentage of correct labels in the training set. Blue, red and green bars indicate the percentage of samples correctly generated by cGANs trained with cross-entropy \cite{GAN}, Hinge \cite{miyato2018spectral} and least-squares \cite{LSGAN} losses, respectively. As it can be observed, when the level of noise $\alpha$ is reasonably below some threshold, the amount of images correctly generated (\ie, correctly classified by the oracle) is always consistently higher than the amount of clean training samples, meaning that the cGAN can ignore noisy labels to some extent.

Several objectives have been proposed for the GAN formulation, which theoretically minimize different divergences between the data distribution $p_{d}(x)$ and the generated one $p_{g}(x)$. For instance, (i) the original GAN, that uses the cross-entropy loss, is proven to minimize the Jensen-Shannon divergence $D_{JS}(p_d||p_g)$ \cite{GAN}; (ii) the least-squares GAN \cite{LSGAN} is proven to minimize the Pearson \cite{Pearson1992} divergence $D_{P}(p_d+p_g||2p_g)$; (iii) a GAN with a Hinge loss is proved to minimize the reverse KL divergence $D_{KL}(p_g||p_d)$ \cite{miyato2018spectral}. In principle, being the KL divergence not symmetric, minimizing $D_{KL}(p_d||p_g)$ would place high probability everywhere the data occurs, while  $D_{KL}(p_g||p_d)$ should enforce low probability wherever the data does \emph{not} occur \cite{Goodfellow-et-al-2016}, thus yielding models more prone to mode collapse \cite{gan_tutorial}. Furthermore, it has been suggested that the Pearson divergence is more resistant to outliers than the KL divergence \cite{PE1,PE2}, and we can interpret samples with noisy label as outliers in the conditional distributions. We are thus interested in understanding how the different objectives, theoretically associated with different divergences, behave in presence of noisy labels

There seem to be some differences between the three objectives: the least-squares GAN \cite{LSGAN} appears to be less resistant to noisy samples. However, since there is no substantial gap between a Hinge GAN and a cross-entropy GAN,  we will only use the latter from now on in this section.

Figure~\ref{fig:noisy_mnist} \textit{(right)} reports the FID scores (Fr\'echet Inception Distance \cite{NIPS2017_fid}) for the same models. The FID is an indirect measure of image quality, accounting for the distance between the training and the generated distributions (the lower, the better). Interestingly, there seem to be no correlation between the amount of noisy labels and the overall quality of the images. 

\begin{figure}[t!]
\centering
\includegraphics[width=0.8\linewidth]{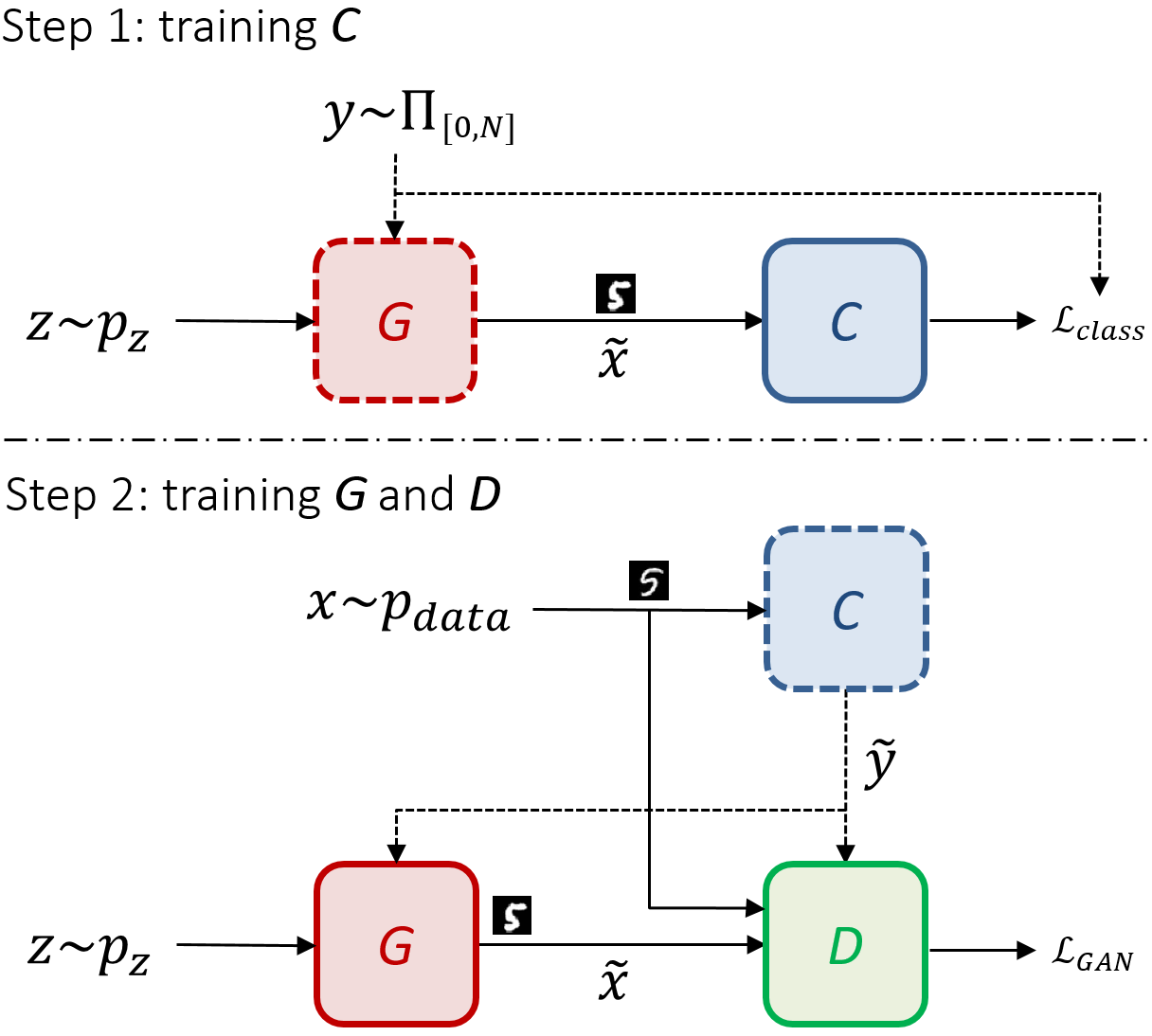}
\vspace{1pt}
\caption{\footnotesize Graphical view of Algorithm~\ref{alg:alg1}. Step 1 \textit{(top)} and step 2 \textit{(bottom)} refer to lines $3-5$ and $6-9$, respectively. The module $G$ and the module $D$ are the generator and the discriminator of the cGAN. The module $C$ is the classifier. Dashed boxes indicate frozen modules (not trained). Solid and dashed wires indicate image and label flows, respectively. $\Pi_{[0,N]}$ is the discrete uniform distribution.}
\label{fig:method}
\end{figure}

\paragraph{Shift Noise.}  As for classifiers, we train several cGANs on $\mathcal{T}=\{x^{(i)}_t,\Tilde{y}^{(i)}\}_{i=1\dots m}$, \ie we try to generate image sets starting from shift-noisy labels. In order to assess model  performances, we exploit the metrics proposed by \cite{Shmelkov_2018_ECCV},  \textit{GAN-test} and \textit{GAN-train}, which specifically deal with classifiers. As already mentioned, the GAN-test is the accuracy of an ``oracle'' classifier trained on real images and evaluated on generated images. This metric tries to capture the precision (i.e., image quality) of GANs. We thus train an oracle classifier for each target set and use it to test the corresponding cGAN trained with shift noise. Ground-truth labels for evaluating the oracle are those fed into the cGAN for class-conditional image generation. The GAN-train is instead the accuracy of a classifier trained on generated images and evaluated on real test images. This metric tries to capture the recall (i.e., diversity) of samples generated \cite{Shmelkov_2018_ECCV}.

Accuracies and asymmetries of such classifiers are reported in Table \ref{tab:delta_acc_GAN}, and confusion matrices are shown in Table \ref{tab:confmats_small}. Interestingly, the samples generated by the CGANs not only induce a better accuracy on the oracle classifier, but also \textit{significantly reduce the amount of asymmetry of the confusion matrices}. This also happens for a classifier trained on generated samples, although to a lower amount.

\paragraph{Summary.}  In conclusion, training a cGAN on the set $\mathcal{T}=\{x^{(i)}_t,\Tilde{y}^{(i)}\}_{i=1\dots m}$ allows to ``filter'' noise in $\Tilde{y}^{(i)}$ in two respects: \textit{i)} by reducing the amount of shift noise in the generated data and \textit{ii)} by reducing the asymmetry of shift noise, making its distribution more alike uniform noise and thus more tolerable for classifiers \cite{Rolnick2017DeepLI}. As a matter of fact, Tables \ref{tab:confmats_small} and \ref{tab:delta_acc_GAN} show that classifiers trained on generated samples (GAN-train column) perform better than classifiers trained with shift noise (Classifier column).

\section{Application to UDA}\label{sec:uda}

In this section, we detail the method designed to face UDA, based on the insights and the findings reported so far.

{\small
\begin{algorithm}[t!]
\caption{Pseudo-Label Refinement (PLR)}
\label{alg:alg1}
\begin{spacing}{1.2}
\begin{algorithmic}[0]
\vspace{3px}
\State \textbf{Input:} target data distribution $p_{target}$, noise distribution $p_{z}$, pre-trained $\theta_D^0, \theta_G^0, \theta_C^0$, step sizes $\eta$, $\delta$
\vspace{3px}
\State \textbf{Output:} learned weights $\theta_D, \theta_G, \theta_C$
\end{algorithmic}
\begin{algorithmic}[1]
\State \textbf{Initialize:} $\theta_D \gets \theta_D^0$, $\theta_G \gets \theta_G^0$, $\theta_C \gets \theta_C^0$
\While{not done}
\State Sample $z \sim p_z$ and $y \sim \Pi_{[0,N]}$
\State Generate $\Tilde{x}=G(z|y)$
\State $\theta_C \gets \theta_C - \eta \nabla_{\theta_C}\mathcal{L}_{class}(\theta_C;\Tilde{x},y)$
\State Sample $x \sim p_{target}$ and $z \sim p_z$
\State Infer $\Tilde{y}=C(x)$
\State $\theta_D \gets \theta_D - \delta \nabla_{\theta_D}\mathcal{L}_{GAN}(\theta_D;z,x,\Tilde{y})$
\State $\theta_G \gets \theta_G - \delta \nabla_{\theta_G}\mathcal{L}_{GAN}(\theta_G;z,\Tilde{y})$
\EndWhile
\end{algorithmic}
\end{spacing}
\end{algorithm}
}

As already mentioned, we can interpret data from the target distribution $p_{target}$, with pseudo-labels (inferred through a classifier trained on data from the source distribution $p_{source}$), as a dataset polluted with label noise. From this perspective, training a cGAN on such empirical, noisy distribution should allow us to generate cleaner samples, as suggested by the findings reported in Section~\ref{sec:robustness}. In turn, a classifier trained on generated data will perform better than the one trained on source data, since noise in the target labels has been reduced in both amount and asymmetry. Starting from these two insights, we define a training procedure where we simultaneously train a classifier $C$ and the modules $G$ and $D$ that define a cGAN (see Figure~\ref{fig:simple}).

\begin{figure*}[!t]
	\begin{center}
		\includegraphics[width=0.9\textwidth]{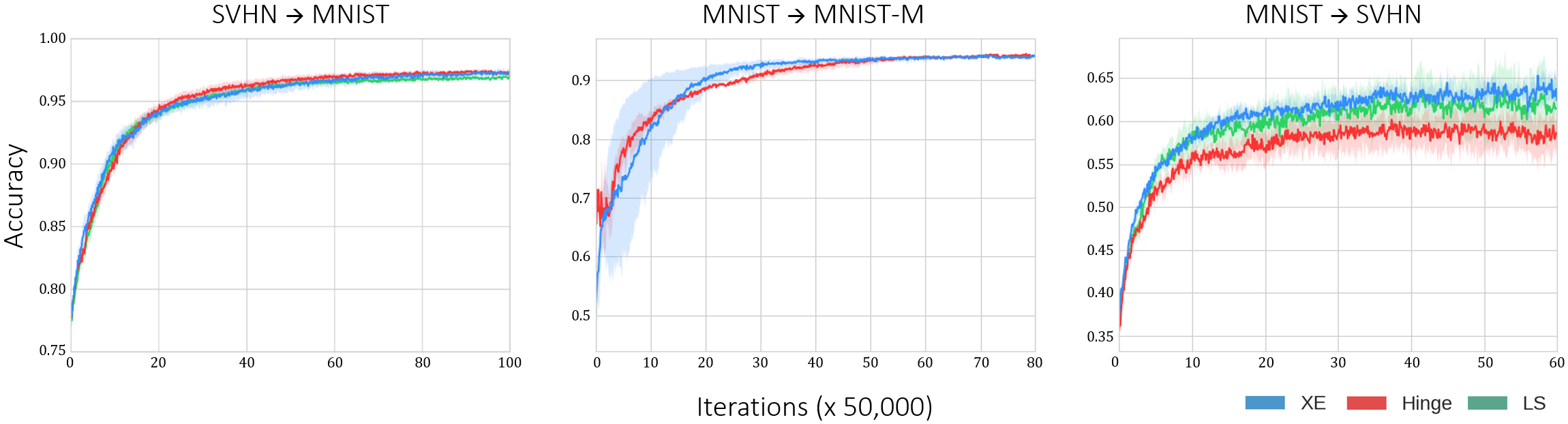}
	\end{center}
	\vspace{-5pt}
	\caption{\footnotesize Evolution of the accuracy on target test sets for SVHN $\rightarrow$ MNIST, MNIST $\rightarrow$ MNIST-M and MNIST $\rightarrow$ SVHN (from \textit{left} to \textit{right}), computed throughout the training procedure described in Algorithm~\ref{alg:alg1}. \textit{Blue}, \textit{red} and \textit{green} curves are associated with GANs trained with the cross-entropy loss \cite{GAN}, the Hinge loss \cite{miyato2018spectral} and the least-squares loss \cite{LSGAN}, respectively. Results obtained with the least-squares loss are not reported for the MNIST $\rightarrow$ MNIST-M as they are significantly worse than the ones achieved with the other options. Curves are averaged over three different runs, shades represent the confidence bands.} 
\label{fig:plots}
\end{figure*}

The pre-train step of our method consists in training a model $M_{\theta_s}$ on labeled data from the source distribution

\begin{equation}
\min_{\theta_{C}} \mathcal{L}_{class} \coloneqq \mathbb{E}_{x,y \sim p_{source}} H(x,y;\theta_s), 
\end{equation}
where $H$ is the cross-entropy loss. Equipped with this classifier, we can straightforwardly infer pseudo-labels for each target sample as $\Tilde{y_t} = C(x_t)$. Typically, an unknown percentage of these labels will be wrong, due to the domain shift between $p_{source}$ and $p_{target}$. We obviously do not know which labels are correct and which are not, but this is irrelevant for the devised strategy. 

Before starting the joint training procedure, we also need to train a cGAN on the noisy target distribution, as in the previous section. This is necessary because we will train $C$ on generated data, and thus starting with randomly-initialized $G$ and $D$ would result in a random classifier $C$, and consequently in non-informative pseudo-labels.

We train the cGAN in a standard fashion, alternating between the following minimax game. Note that we report here the objective as defined in Goodfellow et al. \cite{GAN}, but in our experiments we also test least-squares GANs \cite{LSGAN} and Hinge-GANs \cite{miyato2018spectral}.   

\begin{align}
\min_{\theta_{D}} \max_{\theta_{G}} \, &\mathcal{L}_{GAN} \coloneqq \mathbb{E}_{x,\Tilde{y} \sim p_{target}} [-\log D(x|\Tilde{y})]\\ 
&\quad+  \mathbb{E}_{z \sim p_{z}} [-\log (1 - D(G(z|\Tilde{y})|\Tilde{y}))] \nonumber
\end{align}


\addtolength{\tabcolsep}{3pt}    
\begin{table*}[!h]
\begin{center}
{\footnotesize

\begin{tabular}{@{}rccccc@{}}
\toprule
$\quad$ & \textbf{SVHN $\rightarrow$ MNIST} & \textbf{MNIST $\rightarrow$ SVHN}  &  \textbf{MNIST $\rightarrow$ MNIST-M}  &  \textbf{USPS $\rightarrow$ MNIST} &  \textbf{MNIST $\rightarrow$ USPS} \\
\midrule
Train on source & $0.682$ & $0.314$ & $0.548$  & $0.612$ & $0.783$ \\
\midrule
DANN \cite{Ganin} & $0.739$ & - & $0.767$  & - & - \\
ADDA \cite{ADDA} & $0.760 \pm 0.018$ & - & - & $0.901 \pm 0.008$ & $0.894 \pm 0.002$ \\
DIFA \cite{volpi2018cvpr} & $0.897 \pm 0.020$ & - & - & $0.897 \pm 0.005$ & $0.962 \pm 0.002$ \\
MECA \cite{morerio2018} & $0.952$ & - & - & - & - \\
ATT \cite{Tri} & $0.862$ & $0.528$ & $0.942$ & - & - \\
AD \cite{saito2018adversarial} & $0.950 \pm 0.187$ & - & - & $0.931 \pm 0.127$ & $0.961 \pm 0.029$ \\
MCD \cite{Saito_2018_CVPR} & $0.962 \pm 0.004$ & - & - & $0.941 \pm 0.003$ & $0.965 \pm 0.003$ \\
CoGAN \cite{CoGAN} & - & - & - & $0.931$ \cite{UITITN} & $0.957$ \cite{UITITN} \\
DTN* \cite{DTN} & $0.849$ & - & - & - & - \\
UNIT* \cite{UITITN} & $0.905$ & - & - & $0.936$ &  $0.960$\\
PixelDA** \cite{GOOGLE} & - & - & $0.982$ & - & $0.959$ \\
SBADA** \cite{Russo_2018_CVPR} & $0.761$ & $0.611$ & $0.994$ & $0.950$ & $0.976$ \\ 
GenToAd \cite{GenToAdapt} & $0.924 \pm 0.009$ & - & - & $0.908 \pm 0.013$ & $0.953 \pm 0.007$ \\ 
CycADA \cite{CycADA} & $0.904 \pm 0.004$ & - & - & $0.965 \pm 0.001$ & $0.956 \pm 0.002$ \\
\midrule
\textbf{\emph{Ours (PLR)}} & & & \\
\textit{Cross-entropy} & $0.973 \pm 0.006$ & $0.634 \pm 0.026$ & $0.943 \pm 0.002$ & $0.918 \pm 0.013$ & $0.893 \pm 0.019$ \\
\textit{Least-squares} & $0.969 \pm 0.003$ & $0.618 \pm 0.060$ & - & $0.916 \pm 0.019$ & $0.903 \pm 0.013$ \\
\textit{Hinge} & $0.973 \pm 0.003$ & $0.586 \pm 0.041$ & $0.938 \pm 0.002$ & $0.891 \pm 0.010$ & $0.907 \pm 0.022$ \\
\midrule
Train on target & $0.992$ & $0.913$ & $0.964$ & $0.992$ & $0.999$ \\
\bottomrule

\end{tabular}
} 
\end{center}
\vspace{-5pt}
\caption{Comparison between our method (Pseudo-Label Refinement - PLR) with different GAN objectives and competing algorithms. Test-set accuracies are the results of averaging over $3$ different runs. (*) Uses extra SVHN data ($531,131$ images). (**) Uses $1,000$ target samples for cross-validation.}
\label{tab:res}
\vspace{-10pt}
\end{table*}

\addtolength{\tabcolsep}{-3pt}

Armed with the pre-trained modules $C$, $G$ and $D$, we can start the training procedure, which is defined by Algorithm~\ref{alg:alg1}. In short, we alternate until convergence between (i) updating the weights of the classifier $\theta_c$ via stochastic gradient descent, with labeled target samples uniformly generated via $G$ (lines $3-5$), and (ii) training the weights of the discriminator $\theta_D$ and of the generator $\theta_G$ via stochastic gradient descent, with target samples from $p_{target}$, with pseudo-labels inferred via the classifier $C$ (lines $6-9$). Alternating the two steps will progressively reduce the amount and asymmetry of the initial shift noise in the target  set.

In Figure~\ref{fig:method}, we show the computation flow of the proposed system: \textit{top} (step 1) and \textit{bottom} (step 2) panels represent modules corresponding to lines $3-5$ and $6-9$ of Algorithm~\ref{alg:alg1}, respectively. 

The output of Algorithm~\ref{alg:alg1} is twofold: (i) the trained modules of the cGAN ($G$ and $D$), and (ii) the trained classifier $C$, which is the module finally used to classify target samples. In the next section, we report performances obtained using $C$ in UDA benchmarks.

\section{Experiments}\label{sec:exp}

We test Algorithm~\ref{alg:alg1} on a variety of UDA benchmarks. In every experiment, we run Algorithm~\ref{alg:alg1} until convergence, intended as convergence of the cGAN minimax game, and use accuracy on target dataset test sets (fed to the classifier $C$) as a metric to evaluate our models and compare them with other adaptation approaches.

\textbf{Benchmarks.}
We test our method on the following cross-dataset digit classification problems: SVHN $\leftrightarrow$ MNIST, MNIST $\rightarrow$ MNIST-M and USPS $\leftrightarrow$ MNIST, following protocols on which UDA algorithms based on GANs are tested \cite{CoGAN,DTN,GOOGLE,Russo_2018_CVPR,GenToAdapt}. In order to work with comparable sizes, we resized all images to $32 \times 32$. For each experiment, we use a ConvNet with architecture \textit{conv-pool-conv-pool-fc-fc-softmax}. For the GAN architectures, we draw inspiration from DCGAN \cite{DCGAN}, though considering different objectives (cross-entropy, least-squares, Hinge). All details regarding architectures and training procedures are reported in the Supplementary Material.   

\subsection{Results}
We report in Figure~\ref{fig:plots} the plots showing the evolution of test accuracy in different experiments throughout the training procedure defined by Algorithm~\ref{alg:alg1}. As it can be observed, the performance on target domain of the classifier trained on target samples generated via the cGAN is improved over iterations. It is worth highlighting the monotonic increase of performance: early stopping is not feasible in UDA, thus an unstable algorithm is of scarce utility.

A particularly important result in the one related to the MNIST $\rightarrow$ SVHN split. 
The large gap between the two domains, and the fact that labels are provided for the easier, more biased dataset makes this split particularly difficult to tackle \cite{Ganin,Ganin2}. Our method allows to generate SVHN samples that make the classifier $C$ -- trained on them -- better generalizing to the target distribution, improving performance of $\sim 30\%$ with respect to the baseline. The complete analysis of the obtained results, also in comparison with the state-of-the-art methods, is illustrated in the following.

\textbf{Comparison with other methods.}
Table~\ref{tab:res} compares the proposed method performance (Pseudo-Label Refinement, PLR) with the results obtained by several works in the literature. It is worth to note that, nowadays, research in UDA reached a point where it is difficult to state the superiority of a method over the others. Indeed, Table~\ref{tab:res} shows that there is not a single method that performs better than the others in \emph{every} benchmark. 

First, our method shows performance comparable with the state of the art in the SVHN $\rightarrow$ MNIST split benchmark, significantly outperforming more complex image-to-image translation methods \cite{DTN,UITITN,Russo_2018_CVPR,GenToAdapt} that not only rely on more complicated architectures, but also present a training procedure where the objective is weighted by different hyperparameters (which, as previously mentioned, is a significant drawback in UDA). 

Next, an important result is the one related to MNIST $\rightarrow$ SVHN. As discussed above, this is a rather challenging split, and several methods (\emph{e.g.}, \cite{Ganin,ADDA,volpi2018cvpr,morerio2018,CoGAN,UITITN,DTN}) do not show results on this benchmark. Furthermore, we tested the implementation of PixelDA~\cite{GOOGLE} provided by the authors and could not observe any sign of adaptation. Our algorithm, with the cross-entropy loss as GAN objective, is the best performing method by a statistically significant margin. We also stress that Russo et al. \cite{Russo_2018_CVPR}, the second best performing method on this split, use $1,000$ samples from SVHN to cross-validate the hyperparameters, thus making the working setup much easier than ours. 

On MNIST $\rightarrow$ MNIST-M, the performance achieved with our method is comparable with Saito et al. \cite{Tri} and below the one achieved by methods that perform hyperparameter cross-validation \cite{GOOGLE,Russo_2018_CVPR}. 

\section{Conclusion}\label{sec:concl}
We introduce the concept of shift noise and analyze the robustness of classifiers and cGANs against such highly structured noise. We empirically show that, while classifiers are generally not robust against this kind of label noise, cGANs are more resilient against it, and furthermore generate samples with a more uniform noise distribution. Inspired by these findings, we design a training procedure that progressively allows to generate cleaner samples from the target distributions, and in turn to train better classifiers. 

For future work, we hope to extend the devised algorithm to more realistic UDA benchmarks, such as Office-31~\cite{Saenko2010} and VisDA~\cite{VisDA2017}. The limitation towards this goal is the current computational expense in trainig GANs that generate high-resolution samples~\cite{karras2018progressive,brock2018large}.    

\newpage

\appendix

\section{Architectures and Hyperparameters}

We provide here a detailed description of the networks used for our experiments.

\begin{figure*}
	\begin{center}
		\includegraphics[width=0.9\textwidth]{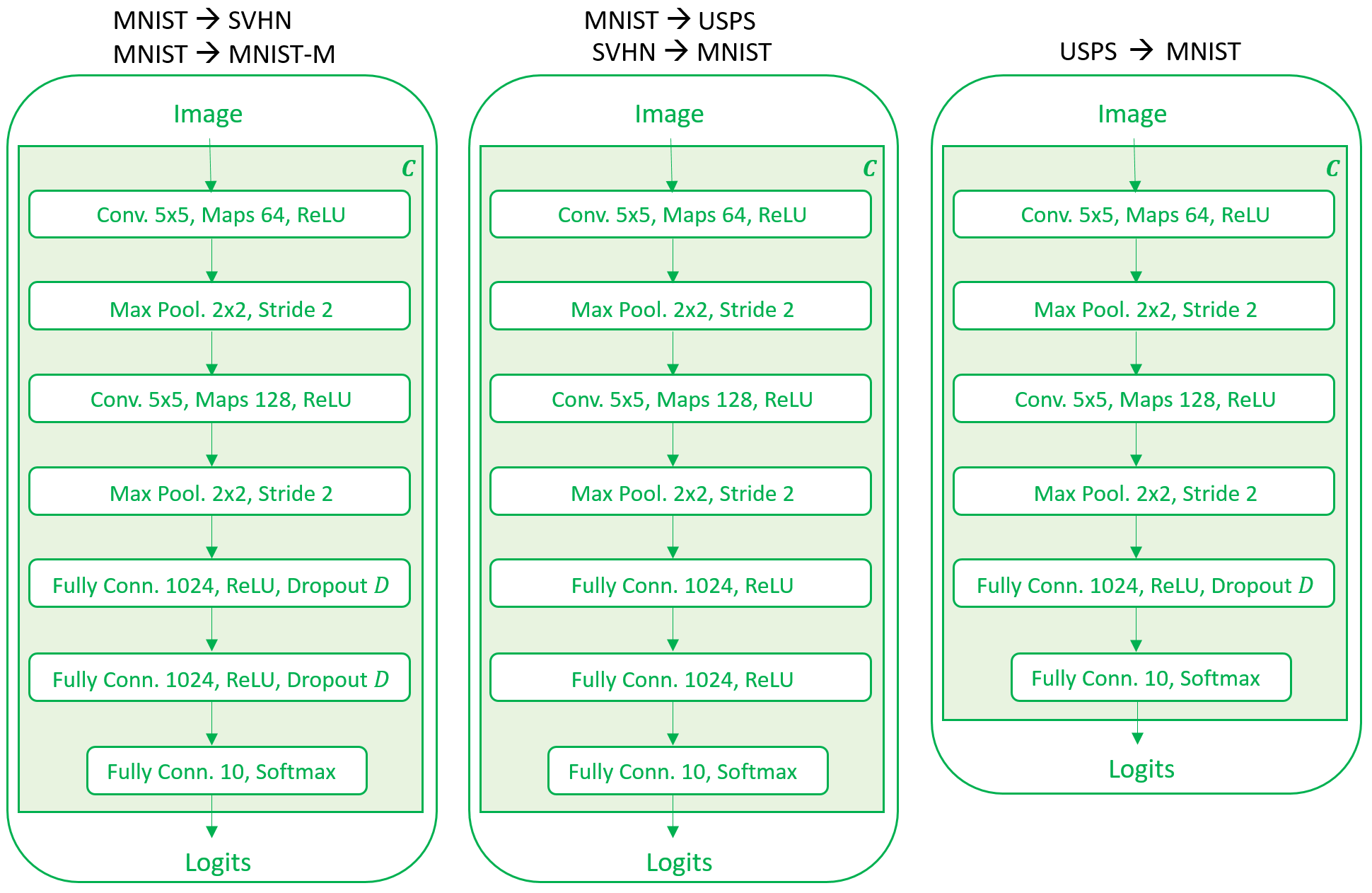}
	\end{center}
	\caption{Architectures for the classifier $C$ (see Figure 4 in the paper).} 
	\label{fig:C}
\end{figure*}

\begin{figure*}
	\begin{center}
		\includegraphics[width=0.9\textwidth]{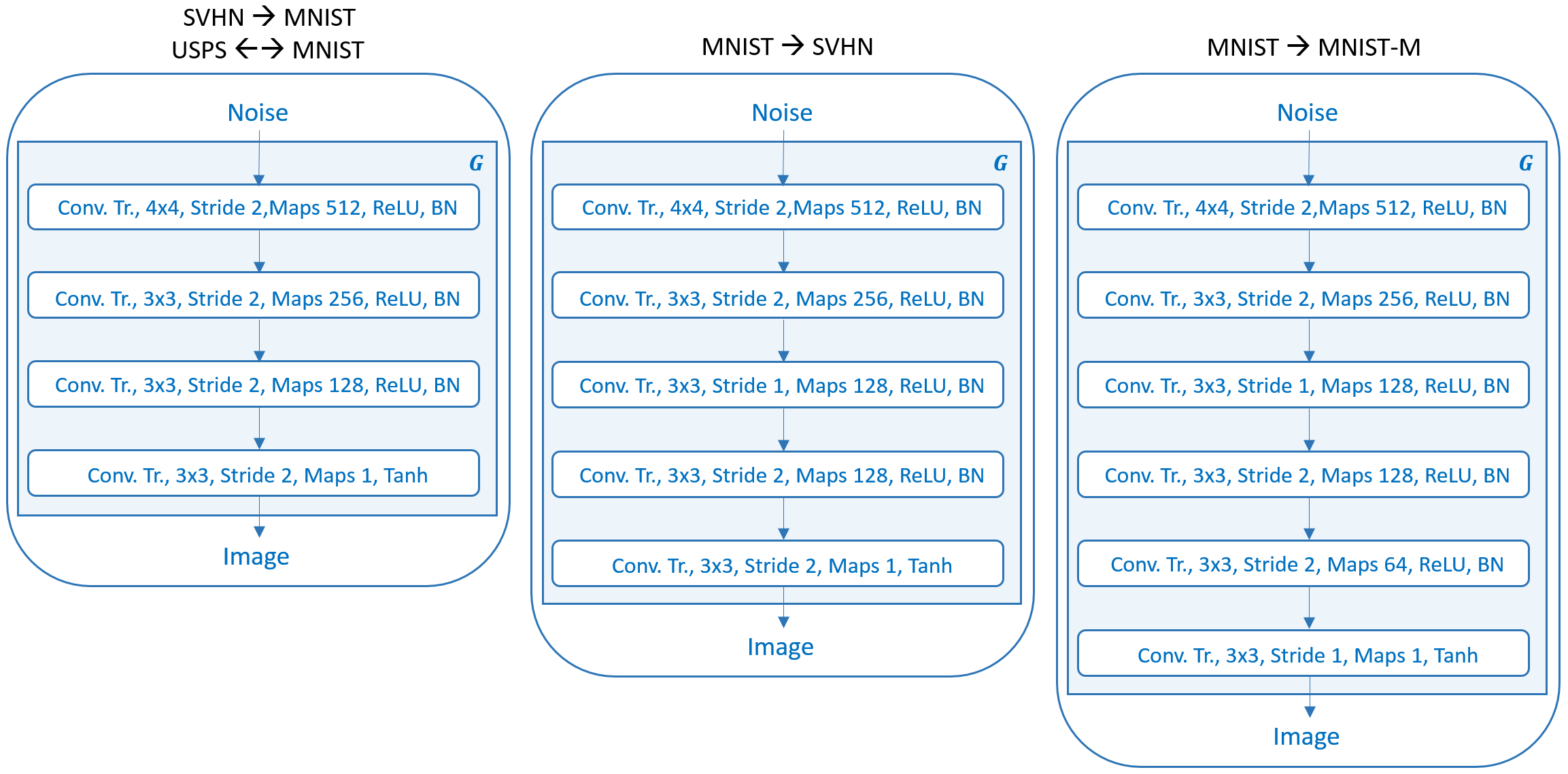}
	\end{center}
	\caption{Architectures for the generator $G$ (see Figure 4 in the paper).} 
	\label{fig:G}
\end{figure*}

\begin{figure*}
	\begin{center}
		\includegraphics[width=0.9\textwidth]{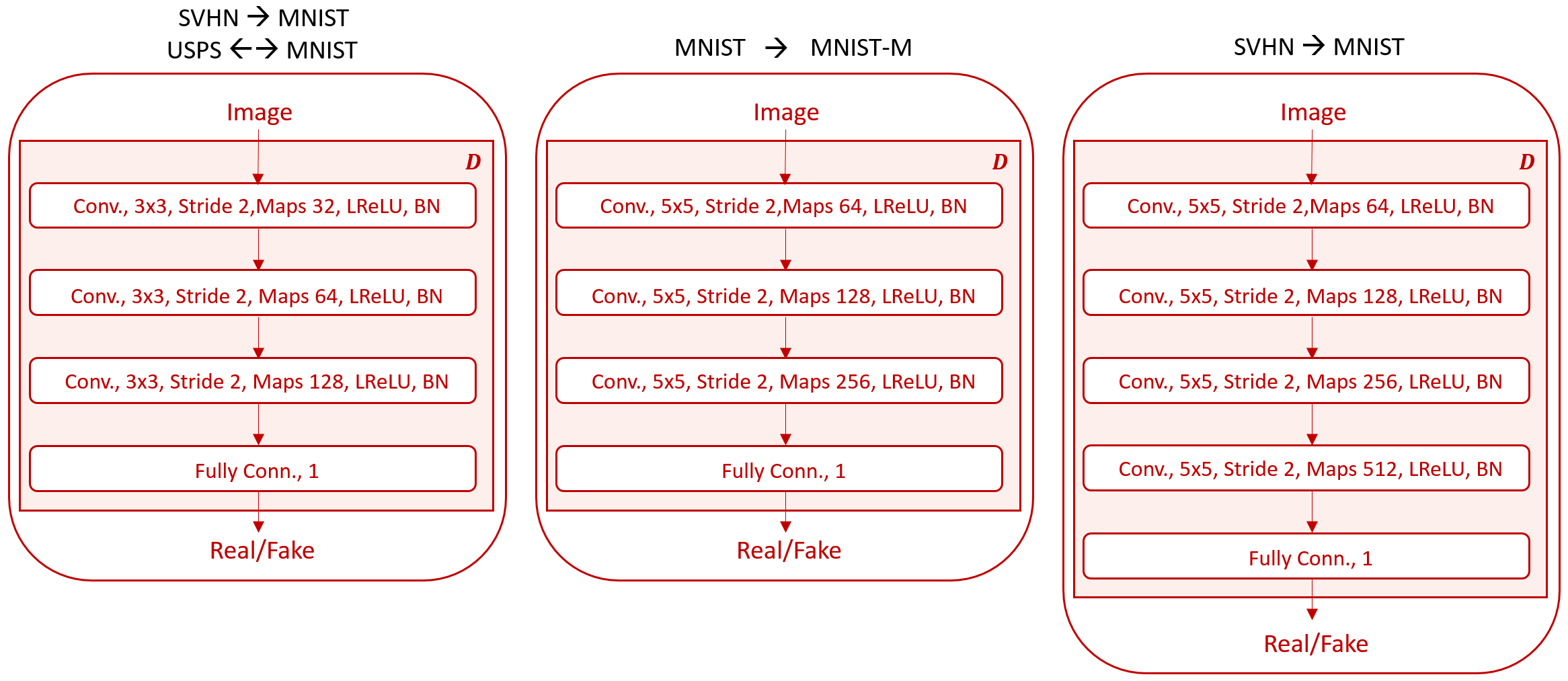}
	\end{center}
	\caption{Architectures for the discriminator $D$ (see Figure 4 in the paper).} 
	\label{fig:D}
\end{figure*}

Figure~\ref{fig:C}, Figure~\ref{fig:G} and Figure~\ref{fig:D} depict the architectures used for $C$ (classifier), $G$ (GAN's generator) and  $D$ (GAN's discriminator), respectively, in the different benchmark experiments.

We report in the following the hyperparameters associated with the same experiments. We use Adam optimizer \cite{ADAM} in all the experiments, and set the learning rate to train pre-train $C$ on data from the source distribution to $3 \cdot 10^{-4}$. For the cGAN pre-training, we set the learning rate for training both $G$ and $D$ to $10^{-5}$. When running Algorithm 1, we set $\eta= 10^{-5}$ and $\delta=5 \cdot 10^{-5}$.

Architectural choices, as well as hyperparameter tuning, were carried out with the goal of making GANs converge.

\section{Are deeper architectures more resistant against shift noise?}

\begin{figure*}
	\begin{center}
		\includegraphics[width=0.6\textwidth]{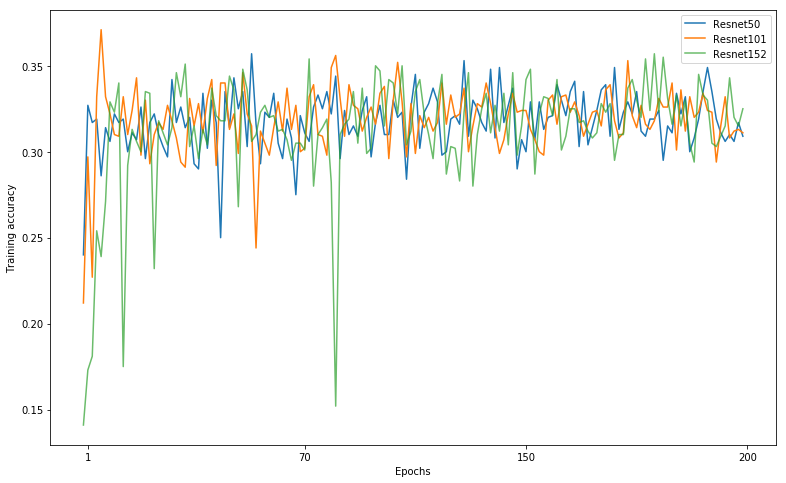}
	\end{center}
	\caption{Training on the shift noise: we evaluate the accuracy on \textit{clean training set} at the end of each epoch. Despite ResNet models are deeper and more resistant to uniform noise \cite{Rolnick2017DeepLI}, they are not robust against shift noise. Indeed, accuracy on the \textit{noisy training set} reaches about 100\% pointing out that the models overfit noise.} 
	\label{fig:resnet}
\end{figure*}

In \cite{Rolnick2017DeepLI} the authors provide empirical evidence that deeper models (e.g. Residual Networks \cite{ResNet}) are more robust against uniform label noise than shallow architectures. We investigated whether such resilience of deep models arises also with shift noise. Our experiments led us to exclude such hypothesis. We considered the split MNIST $\rightarrow$ SVHN, where shift noise is very significant, and repeated the experiment of Table 2: we trained different ResNets (from scratch) with different depths on target samples corrupted by shift noise; we observe that despite improved capacity of the models, they overfit the noisy labelled samples and are not able to reduce $\delta_A$. (see Table \ref{tab:delta_acc_classif_supp} and Figure \ref{fig:resnet}).

\begin{table*}[h]
\begin{center}
\small{
\begin{tabular}{r|cccc}
\toprule
\multirow{2}{*}{\textbf{Architecture}} & \multicolumn{2}{c}{\textbf{Shift noise}} &  \multicolumn{2}{c}{\textbf{Classifier}} \\

\cmidrule(lr){2-3}  \cmidrule(lr){4-5} 

 & $a$ & $\delta_A$ & $a$ & $\delta_A$ \\
\midrule

$C$
& \multirow{4}{*}{0.3005} & \multirow{4}{*}{0.3739}
& 0.3212	 & 0.3741 \\

ResNet-50
& & 
& 0.2998 & 0.3741\\

ResNet-101
& & 
& 0.3004 & 0.3739\\

ResNet-152
& & 
& 0.2997 & 0.3736\\

\bottomrule
\end{tabular}
} 
\end{center}
\caption{MNIST $\rightarrow$ SVHN: we observe no improvements in accuracy wrt shallower classifiers that we trained in the paper. Even $\delta_A$'s do not sink as it happens for generative models.}
\label{tab:delta_acc_classif_supp}
\end{table*}



\section{Generated images}

\begin{figure*}
	\begin{center}
		\includegraphics[width=0.6\textwidth]{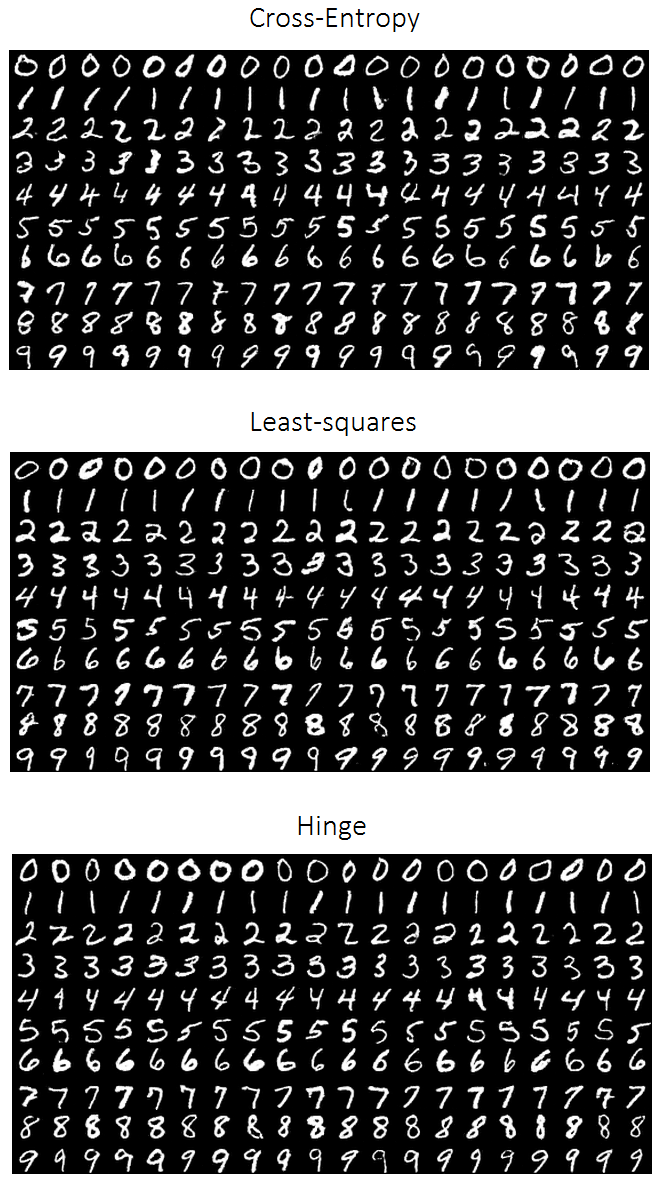}
	\end{center}
	\caption{MNIST samples generated by $G$, trained with Algorithm 1 (SVHN $\rightarrow$ MNIST split). Each row is related to a different label code (from \emph{top} to \emph{bottom}, $0$ to $9$).} 
	\label{fig:s_mnist}
\end{figure*}

\begin{figure*}
	\begin{center}
		\includegraphics[width=0.6\textwidth]{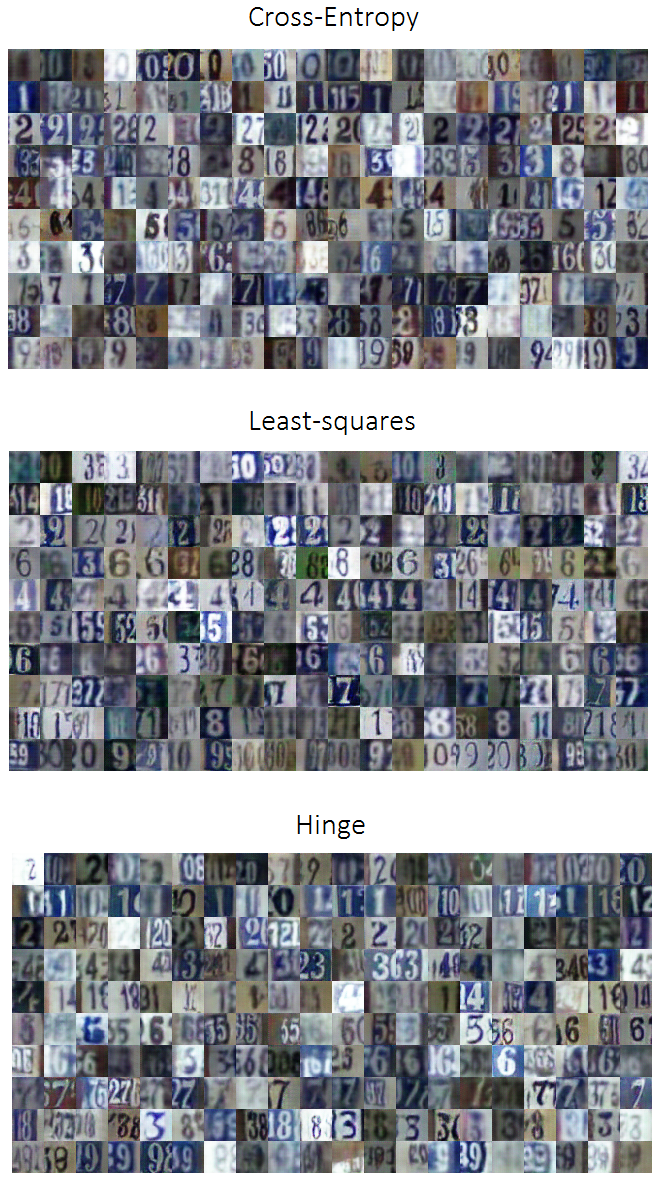}
	\end{center}
	\caption{SVHN samples generated by $G$, trained with Algorithm 1 (MNIST $\rightarrow$ SVHN split). Each row is related to a different label code (from \emph{top} to \emph{bottom}, $0$ to $9$).} 
	\label{fig:svhn}
\end{figure*}

\begin{figure*}
	\begin{center}
		\includegraphics[width=0.6\textwidth]{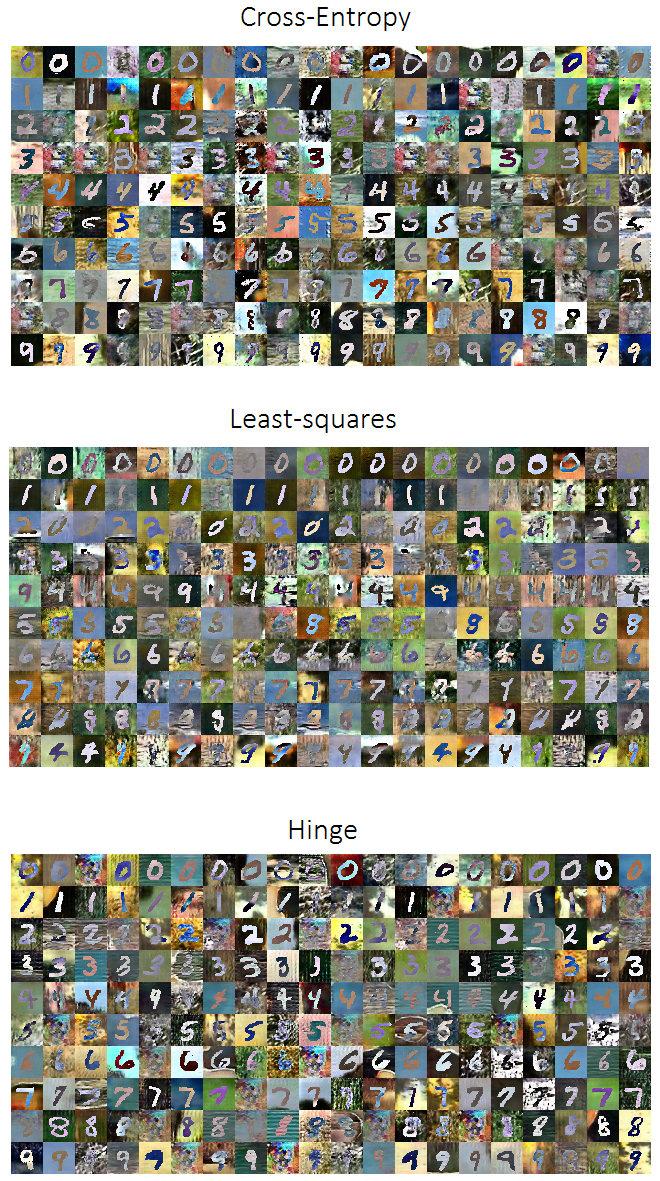}
	\end{center}
	\caption{MNIST-M samples generated by $G$, trained with Algorithm 1 (MNIST $\rightarrow$ MNIST-M split). Each row is related to a different label code (from \emph{top} to \emph{bottom}, $0$ to $9$).} 
	\label{fig:mnist_m}
\end{figure*}

\begin{figure*}
	\begin{center}
		\includegraphics[width=0.6\textwidth]{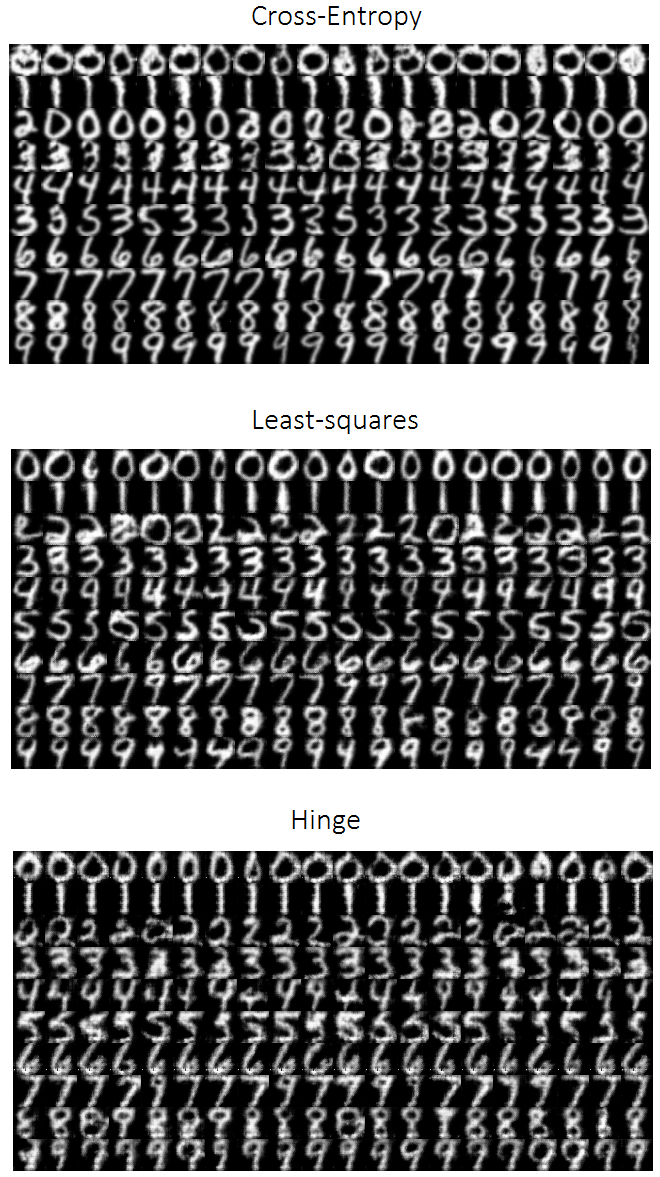}
	\end{center}
	\caption{USPS samples generated by $G$, trained with Algorithm 1 (MNIST $\rightarrow$ USPS split). Each row is related to a different label code (from \emph{top} to \emph{bottom}, $0$ to $9$).} 
	\label{fig:usps}
\end{figure*}

We report in Figures~\ref{fig:s_mnist}, \ref{fig:svhn}, \ref{fig:mnist_m}, \ref{fig:usps} and \ref{fig:u_mnist} samples generated by $G$ after the training procedure defined by Algorithm 1, for the splits SVHN $\rightarrow$ MNIST, MNIST $\rightarrow$ SVHN, MNIST $\rightarrow$ MNIST-M, MNIST $\rightarrow$ USPS and USPS $\rightarrow$ MNIST, respectively. For each experiment, we randomly generated $20$ samples associated with the different classes and reported them in the Figures, where each row is related to a different class.

\clearpage

{\small
\bibliographystyle{ieee}
\bibliography{egbib}
}
\end{document}